# Supervised Saliency Map Driven Segmentation of Lesions in Dermoscopic Images

Mostafa Jahanifar*, *Student Member, IEEE*, Neda Zamani Tajeddin*, *Student Member, IEEE*, Babak Mohammadzadeh Asl, *Member, IEEE,* and Ali Gooya, *Member, IEEE*

*Abstract*—**Lesion segmentation is the first step in most automatic melanoma recognition systems. Deficiencies and difficulties in dermoscopic images such as color inconstancy, hair occlusion, dark corners and color charts make lesion segmentation an intricate task. In order to detect the lesion in the presence of these problems, we propose a supervised saliency detection method tailored for dermoscopic images based on the discriminative regional feature integration (DRFI). DRFI method incorporates multi-level segmentation, regional contrast, property, background descriptors, and a random forest regressor to create saliency scores for each region in the image. In our improved saliency detection method, mDRFI, we have added some new features to regional property descriptors. Also, in order to achieve more robust regional background descriptors, a thresholding algorithm is proposed to obtain a new pseudo-background region. Findings reveal that mDRFI is superior to DRFI in detecting the lesion as the salient object in dermoscopic images. The proposed overall lesion segmentation framework uses detected saliency map to construct an initial mask of the lesion through thresholding and post-processing operations. The initial mask is then evolving in a level set framework to fit better on the lesion's boundaries. The results of evaluation tests on three public datasets show that our proposed segmentation method outperforms the other conventional state-of-the-art segmentation algorithms and its performance is comparable with most recent approaches that are based on deep convolutional neural networks.**

## I. INTRODUCTION

NON-MELANOMA and melanoma skin cancers are both increasing in the recent decades. World Health Organization (WHO) reports that melanoma globally occurrence is about 132,000 each year, and by depleting the ozone this statistics will be increased [1]. Fortunately, a high survival rate is reported for melanoma as long as it is diagnosed in early stages [2]. Nowadays, computers and smart

Mostafa Jahanifar was with the Tarbiat Modares University, Tehran, Iran. He is now with the Department of Research and Development, NRP company, Tehran, Iran (e-mail: m.jahanifar@modares.ac.ir).

Neda Zamani Tajeddin and Babak Mohammadzadeh Asl are with the Biomedical Engineering Department, Tarbiat Modares University, Tehran, Iran (e-mails: {n.zamanitajeddin, babakmasl}@modares.ac.ir).

Ali Gooya is with the Department of Electronic and Electrical Engineering, University of Sheffield, Sheffield, England (e-mail: a.gooya@sheffield.ac.uk).

*These authors contributed equally to this work.

hand-held devices are very popular and can help to diagnose melanoma earlier. Computer Aided Diagnosis (CAD) tools can be coupled with these devices to construct an intelligent system, which is able to help dermatologists in melanoma recognition. In fact, in a recent research, Codella *et al.* [3] showed that some of these CAD systems are able to perform better than an average agreement of human experts.

Classical CAD systems for melanoma recognition usually consist of three main parts: lesion segmentation, feature extraction, and feature classification [4]. Yu *et al.* [5] have shown that although the classification of melanoma patients can be performed using only the features extracted by deep learning models, the diagnostic performance is improved significantly by incorporating segmentation of the lesions.

Different methods have been proposed in the literature for melanoma segmentation. Silveira *et al.* [6] assessed six different algorithms that were based on adaptive thresholding, active contours, levels sets, or regional information. A lot of researchers segmented the lesion using thresholding methods [7], [8]. For example, Barata *et al.* [7] analyzed the histogram in order to obtain a suitable threshold. In some methods, only regional information was adopted and the lesion was segmented by clustering or splitting the regions [9]–[11] e.g., Pennisi *et al.* [11] applied Delaunay triangulation to partition the image into several regions and segment the lesion by merging them. Combinations of active contour models with other methods were also exploited several times [4], [6], [12]. In our previous work [4] a histogram thresholding algorithm was used to achieve initial border of the lesion, which was then propagated towards the actual lesion boundary, using a dual-component speed function in a level set framework. Some efforts have been made to segment the lesion by mimicking the way that dermatologists delineate the lesion boundaries [13], [14]. Recently, convolutional neural networks proved to perform very well on dermoscopic image segmentation [5], [15]. In order to review the skin lesion segmentation methods more comprehensively, the most recent surveys on this topic are suggested [16], [17].

An acceptable lesion segmentation algorithm must be able to handle the present deficiencies in dermoscopic images. Deficiencies like color inconstancy, hair occlusion, indistinct lesion borders, the presence of ruler marks, dark corners, color charts, and marker inks make difficulties in proposing a general lesion segmentation algorithm [18]. Most algorithms overcome difficulties by incorporating preprocessing techniques [4], [18]. However, a powerful segmentation algorithm must show little sensitivity to these problems.

In this paper, we propose a hybrid framework for lesion segmentation, comprising three main parts: preprocessing, initial mask creation, and final mask creation. The initial mask of the lesion is constructed by thresholding the saliency map of the image and the final mask is obtained through a distance regularized level set method. The contribution of this work mostly relies on the construction of the saliency map. Based on the prior information from dermoscopic images, we propose to integrate some new features into a well-known supervised saliency detection framework in order to boost its performance on the lesion detection task and handle the difficulties in dermoscopic images.

After reviewing the related works based on saliency detection, the rest of paper is organized as follows: in section II the supervised saliency detection algorithm is described. Our proposed segmentation framework, modified saliency detection algorithm and its novelties are thoroughly explained in section III. Results of applying proposed methods on three different datasets and discussion about their properties are presented in sections IV and V. Finally, the paper is concluded in section VI.

### A. Related works based on saliency detection

The salient object can be heuristically defined as the most prominent object in the image, the region of the image that is noticed at first sight, or the segment of the image that has the most contrast from the background [19]. Salient object detection has been an active field of research in the recent years and there are many articles published on this topic [20]. Saliency object detection methods can be divided into two groups: unsupervised and supervised saliency detection mechanisms [20]. In the unsupervised variant, saliency map is created directly from image information and features. Most of these methods characterize image contrast in different regions to obtain the saliency map. For the supervised variant, like the approach used in this study, several features are extracted from image regions and a predictor model, trained on a labeled dataset, constructs the saliency map [21].

Lesion segmentation through saliency detection approaches has been addressed in the recent years [22], [23]. Fan *et al.* [22] proposed an unsupervised approach to construct two saliency scores, one from image color information and another from image brightness information, and then create final saliency map by combining these two saliency scores. Lesion segmentation is then achieved by thresholding the saliency map through a histogram analyzing method [22]. They assumed that marginal area of the image belongs to healthy skin, which does not necessarily hold. They emphasized on contrast property of lesion in images, however, some lesions may have not enough contrast to fulfill their criteria. Besides, color charts show a high contrast in images and may be incorrectly detected as a lesion or a salient object.

Ahn *et al.* [23] first segmented images into several superpixels and found background regions using a multi-scale framework. After identifying background regions they proposed to create the saliency map via sparse reconstruction error [23]. In other words, they claimed that the regions with larger reconstruction error are more likely to belong to the lesion. In the rest of their method, they incorporate several saliency map refinements and a thresholding algorithm. Even though Ahn *et al.* [23] addressed the problem of lesions reaching to the image boundary and achieved better results in comparison to Fan *et al.* study [22], they did not deal with color charts and dark corners in the images. Thus, they simply removed these deficiencies manually before any processing, which is a tedious work.

In this paper, we propose a saliency detection method specially tailored for dermoscopic images which, not only uses contrast and background descriptors, but also takes the location, shape, color, and texture information of image regions into account to achieve a better result on saliency detection. Our saliency detection method is able to automatically handle deficiencies like hair occlusion, ruler marks, color charts, dark corners, and the problem of lesions reaching to image borders. The general ideas of the supervised saliency detection framework are described in section II, but our incorporated novelties in that framework are explained in section III.

## II. SUPERVISED SALIENCY DETECTION

In this paper, like the work of Wang *et al.* [21], a multi-level saliency detection scheme based on discriminative regional feature integration (DRFI) is used. DRFI is reported to be one of the most efficient algorithms for saliency detection based on recent benchmarks [19], [20].

A multi-level approach has been implemented in DRFI framework. The three main steps of the Wang *et al.*'s DRFI algorithm [21] are listed below:

1. Multi-level segmentation: decomposition of the image to its constituent elements from a fine level to a coarse level.

2. Saliency regression: using a random forest regressor to map the regional feature vector into a saliency score.

3. Saliency map fusion: creating the final saliency map by fusing saliency maps obtained from different levels.

A diagram of implemented saliency map construction framework and the outputs of its various steps are illustrated in Fig. 1. Each of the abovementioned steps has its own consideration, which we will explain briefly in the following subsections.

### A. Multi-level segmentation

According to [21], in multi-level segmentation, a graph-based image partitioning algorithm is used [24]. The first level of segmentation (the finest) has the most number of output regions. For next levels, the segmentation is done by merging adjacent similar regions in their previous level. Thus, the final level of segmentation (the coarsest) is likely to have the least number of regions. It is important to note that we are taking the relation of neighboring regions into account for saliency regression by working in a multi-level segmentation framework.

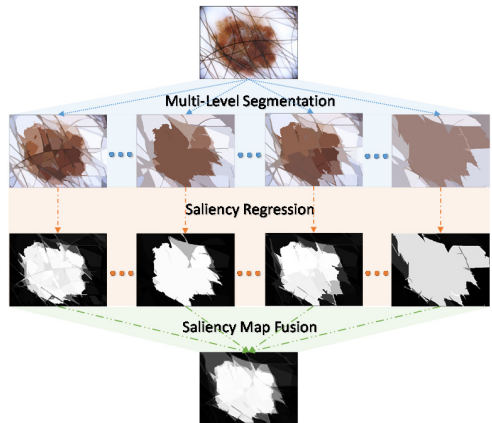

Fig. 1. DRFI saliency map detection procedure.

## TABLE I
### LIST OF REGIONAL CONTRAST AND BACKGROUND DESCRIPTORS

| features | dimension |
|---|---|
| Average RGB values differences | 3 |
| Average La*b* values differences | 3 |
| Average HSV values | 3 |
| Absolute response of the Leung-Malik filters differences | 15 |
| Max response among the Leung-Malik filters difference | 1 |
| La*b* histogram difference | 1 |
| Hue histogram difference | 1 |
| Saturation histogram difference | 1 |
| Local Binary Pattern histogram difference | 1 |

## TABLE II
### LIST OF REGIONAL PROPERTY DESCRIPTORS

| features | dimension |
|---|---|
| Average normalized x and y coordinates | 2 |
| $10^{th}$ and $90^{th}$ percentiles of normalized x and y coordinates | 4 |
| Normalized area and perimeter | 2 |
| Normalized area of the neighboring regions | 1 |
| Aspect ratio of the bounding box | 1 |
| Variance of the RGB values | 3 |
| Variance of the La*b* values | 3 |
| Variance of the HSV values | 3 |
| Variance of the response of the LM filters | 15 |
| Variance of LBP feature | 1 |
| **Average of the RGB values** | **3** |
| **Average of the a* and b* values (from La*b*)** | **2** |
| **Shape elongation** | **1** |
| **Shape extent** | **1** |
| **Circle probability** | **1** |
| **Energy of Laws' filters responses** | **14** |
| **Segmentation level** | **1** |

Bold faced items are newly introduced in this work.

## B. Saliency regression for each level

In the second step of DRFI, regional features are extracted from each segmented region (at different levels) of the image. There are three types of regional features in DRFI approach [21]: regional contrast, property, and background descriptors.

### 1) Regional contrast descriptors

Wang *et al.* [21] proposed to extract features representing the regional contrast. To compute the regional contrast descriptors, features from each region are compared with those extracted from their neighboring areas.

For a specific segmentation level, different features, denoted by $v^R$ vectors, will be extracted from its regions. In addition, considering all adjacent regions as a single neighboring region, we can extract the same features and construct the neighborhood feature vector $v^N$. To measure each region contrast descriptor, DRFI method proposes to calculate the difference of $v^R$ and $v^N$ as follows:

$$d(v^R, v^N) = \begin{cases} \sum_{i=1}^{b} \dfrac{2(v_i^R - v_i^N)^2}{v_i^R + v_i^N}; & \text{histogram} \\ & \text{inputs} \\ (|v_1^R - v_1^N|, ..., |v_b^R - v_b^N|); & \text{O.W.} \end{cases} \quad (1)$$

meaning that for the features that comprise a histogram, a normalized sum of differences in histogram bins is considered as the feature difference, and for the other forms (regional features that are not histogram), the absolute elementwise differences are computed. In (1), *b* refers to the number of dimensions of the feature vectors. Wang's proposed feature set for contrast description is summarized in Table I, for further information refer to [21].

### 2) Regional property descriptors

This group of features directly describe the shape, location, color, or texture of a region. These features can be very useful when contrast descriptors are inefficient. For example, in a dermoscopic image, a color chart indicator with green color shows a great contrast against the healthy skin, whereas it is not an eligible salient object that we are seeking. Thus, the contrast descriptor alone cannot work very well in that situation. Fortunately, the regressor model in DRFI framework can easily be trained to neglect green regions in the image by adding some regional descriptors in the training phase. Wang *et al.* [21] proposed 35 regional property features for DRFI listed in Table II (with regular font).

### 3) Regional background descriptors

It is shown that region properties cannot identify the background in natural images [19], [21]. By assuming this hypothesis that most of the salient objects are placed in the center of the image, DRFI considers a pseudo-background region around the image, in order to calculate the degree of region's belonging to the background. In [21], a border of 15 pixels around the image area was picked out as the pseudo-background region. Just like regional contrast descriptors, DRFI calculates the difference of each region's features from the features of the pseudo-background region, to serve as the regional background descriptors.

### 4) Saliency regression

By combining the regional contrast, property, and background descriptors, a regional feature vector is formed to describe the regions. For each level in the multi-level segmentation framework, regional feature vectors should be extracted from all regions of that level. The corresponding labels (whether a region belongs to the salient object: 1 or not: 0) for each region are also available through their ground truth

Copyright (c) 2018 IEEE. Personal use is permitted. For any other purposes, permission must be obtained from the IEEE by emailing pubs-permissions@ieee.org.

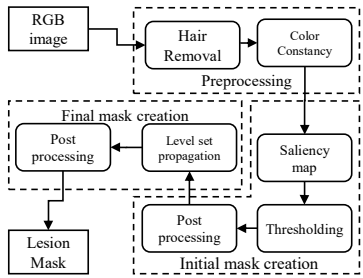

Fig. 2.  An overview diagram of the proposed segmentation framework.

segmentation mask. Regional feature and label vectors are then utilized to train a random forest regressor [25]. In the prediction phase, the trained regressor gets the saliency feature vector of a region and returns a saliency score between 0 and 1 for it. By predicting the saliency scores for all regions in an image (for a certain level), its saliency map is constructed.

### C.  Saliency map fusion

To achieve the final saliency map, saliency maps of different levels must be fused. A linear combinator is proposed for this task in DRFI framework [21]. This combinator constructs the final saliency map by operating a weighted integration of multi-level saliency maps. DRFI learns the combination weights through the least square method on the training images [21].

To the best of our knowledge, the current work is the first research that uses such supervised approach for detecting lesions as salient objects in dermoscopic images. In order to improve the detection of skin lesions in DRFI framework, we have added 5 color, 3 shape, and 14 texture-related features to the regional property descriptors, which will be described in the next section. Also, we have introduced a new pseudo-background region to better distinguish between the lesion and the background. Apart from these, some new thresholding, contour evolution, and postprocessing techniques are used to accurately segment the lesion.

### III.  Proposed Segmentation Method

For segmenting the lesion in dermoscopic images, we follow the method illustrated in the diagram of the Fig. 2. Proposed segmentation method consists of three main parts: preprocessing, initial mask creation, and final mask creation. The preprocessing steps aim to compensate for the deficiencies in the image. The goal of the second part of the algorithm is to construct an initial mask for the lesion. This mask will further be used in the third part of the algorithm to better delineate lesion border in the image. Each of these parts consists of various algorithmic steps, which will be described in the next subsections.

### A.  Preprocessing

Due to different deficiencies that usually exist in dermoscopic images, preprocessing is a vital task before any further analysis. The most problematic issues in images can be hair occlusion, uneven illumination, color inconstancy, ruler marks, color charts, and dark corners [18]. Lesion segmentation could undergo some difficulties because of these deficiencies. There are a vast variety of methods proposed to solve each of these problems [17]. We are planning to detect the lesion in a supervised learning framework, so we suppose that such algorithms should be able to distinguish color charts or dark corners from the lesion. Therefore, unlike the approaches we took in our earlier work [4], here we only explore the effects of hair removal and color constancy on the image segmentation.

#### 1)  Color constancy

A general dataset of dermoscopic images, like the one in the ISBI 2017 challenge [26], comprises of images captured in different lighting situations using different dermatoscope devices. Based on this fact, image color would change from image to image which lowers the performance of learning algorithms [27]. Here, we propose to use color constancy to reduce color variation in dataset images using Shades of Gray algorithm [27]. Shades of Gray estimates the color of illuminant in the image based on each channel weighted norm $e_c$ calculated using (2), in which $I_c$ is one of the image channels $c \in \{R, G, B\}$ in the image domain $\mathbf{x}$, $k$ is a normalization constant and $p$ is the norm degree [27]:

$$e_c = \frac{1}{k} \left( \frac{\int (I_c(\mathbf{x}))^p \, d\mathbf{x}}{\int d\mathbf{x}} \right)^{1/p}. \tag{2}$$

Based on the above equation, the normalization constant $k$ can be computed by setting $\sqrt{e_R^2 + e_G^2 + e_B^2} = 1$. Finally, a correction coefficient for each channel is calculated as $d_c = (\sqrt{3}e_c)^{-1}$. By multiplying each correction coefficient in its corresponding image channel, a color corrected image can be obtained. As in the work of Barata et al. [27], the parameter $p$ is set to 6 in this research. The result of applying such algorithm on a sample dermoscopic image, that is occluded with hairs and has a shade of blue in its illuminant color, is illustrated in Fig. 3. Applying the color constancy seems to improve the contrast of the lesion in comparison to the surrounding skin and the range of colors becomes more consistent with the other images in the dataset.

#### 2)  Hair and ruler marks inpainting

Hair occlusion can jeopardize the segmentation task. In this paper, we use the method described in [28] to detect hair-like structures in the image and replace them with appropriate pixel values. Koehoorn et al. [28] proposed a multi-threshold scheme to initially segment the hairs in the image. In every thresholding step, they used a gap-detection algorithm to detect potentially hair pixels. All initial results are then combined into a single mask. This mask contains some objects that are falsely segmented as hair. Thus, in the last step of algorithm authors used a combination of morphological filters and medial descriptors to validate the hair objects. Detected hair pixels are then replaced using a fast marching based image inpainting algorithm [28]. This approach is capable of detecting and inpainting dark and light hairs from the image.

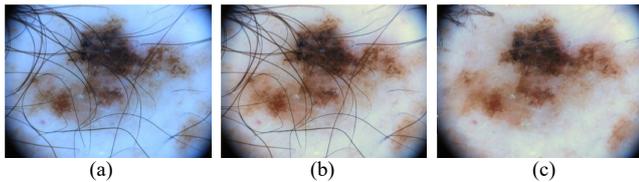

Fig. 3. Image preprocessing procedures applied to a sample dermoscopic image: (a) original image occluded with hairs and showing a shade of blue color, (b) applying Shade of Gray algorithm to achieve color constancy, and (c) removing hair from the image using Koehoorn's method [28].

### B. Initial mask creation through saliency detection

Considering the lesion as the salient object in dermoscopic images, we propose to use the supervised saliency detection framework explained in section II for lesion detection. DRFI approach originally introduced by Wang *et al.* [21] was designed to detect salient objects in a dataset of natural images, which are very challenging. Dermoscopic images, on the other hand, have their own challenges that must be considered in the saliency detection pipeline. Knowing specific properties about dermoscopic images help us extend DRFI feature set to better describe them. Accordingly, in this section, we propose to improve the DRFI approach by incorporating new regional property descriptors and a more relevant pseudo-background region in its framework. This new modified version of DRFI is the key contribution of this paper and we abbreviate it as "mDRFI" in the rest of paper. Apart from these novelties, all other procedures for saliency detection are followed as described in section II.

#### 1) Modifying DRFI for dermoscopic saliency detection

##### 1.1) Extending regional property descriptors

It is shown that the regional property descriptors have a great impact on the saliency regression (based on experiments on natural image datasets, 9 features out of 20 most important regional features belong to this group) [21]. We propose to add the following features to the regional property descriptors of mDRFI in order to improve its functionality for detecting skin lesions as the salient object. Newly added features can be categorized in color, shape, or texture related features.

*Color related features:* in the original DRFI approach, the variance of region pixel values from different color channels (R, G, B, L, a*, b*, H, S, and V) are used as the regional property descriptor. The variance of pixel values is a texture descriptor which measures changes in a region appearance. To capture color properties of a region, we propose to average its values in R, G, B, a*, and b* channels as the absolute color descriptors (5 new features).

*Shape related features:* some structures in the image can be identified via their shapes. Hairs in dermoscopic images are tubular structures which usually have long and thin shapes. We are able to characterize this property of a region by adding a feature that measures the shape elongation:

$$Elongation = 1 - m/M \ , \tag{3}$$

where $m$ and $M$ are respectively the lengths of the minor and major axes of the ellipse that has the same normalized second central moments as the region. We also propose to add a feature measuring the extent of the region to describe rectangularity of its shape. The extent is calculated by dividing the region area by the area of its minimum-area bounding box.

It is very important not to confuse the color charts around some images as the salient object. Some of these charts can easily be recognized by their color (e.g., pure green color chart), but it is common for color charts to have similar colors to lesions. Ergo, it is vital to use their shape and position properties to recognize them and preventing regressor to wrongly detect them as the salient object. For this end, we employ the method of Basalmah [29] for partial circle detection.

The edge map, $\Omega$, of the image is constructed by thresholding the Prewitt gradient of the image. Then, for each pixel of the image, $(x, y)$, the Euclidean distance to every edge pixel, $(i, j) \in \{\Omega = 1\}$, in the edge map is collected in a distance vector, $\mathbf{dist}_{(x,y)} = \{ \text{dist}_{(x,y)}^{(i,j)}, (i,j) \in \{\Omega = 1\} \}$, in which distance function is defined as Euclidean distance:

$$\text{dist}_{(x,y)}^{(i,j)} = \sqrt{(i-x)^2 + (j-y)^2} . \tag{4}$$

The histogram of that distance vector for each pixel location, $\mathbf{H}_{(x,y)}$, is constructed to be used in finding the potential circles and further analysis:

$$
\mathbf{H}_{(x,y)} = \left\{ h_{(x,y)}^d ; d = 0, ..., D \right\}, \text{ in which:}
$$
$$
h_{(x,y)}^d = \left\{ \#\{(i,j) \in \Omega\} \exists \, \text{dist}_{(x,y)}^{(i,j)} = d \right\}. \tag{5}
$$

We form a *centerMap* and a *radiusMap* with the same size as the original image and for each pixel, the maximum values of its histogram are recorded in that pixel position in the *centerMap*. The location of histogram maximum is also recorded in the same pixel position of *radiusMap*:

$$
centerMap(x,y) = \max \left( \mathbf{H}_{(x,y)} \right),
$$
$$
radiusMap(x,y) = \operatorname*{argmax}_{d} \left( \mathbf{H}_{(x,y)} \right). \tag{6}
$$

Dominant peaks (peaks greater than a threshold) indicate that the pixels under investigation can be the center of a circle because they have a great number of edge pixels distributed with equal distance around them. Thus, we convert the *centerMap* to a binary mask which indicates the location of circles' center by applying a threshold on it. Values of the *radiusMap* in the location of centers also indicate their corresponding circle radius. Having the circle's center position and its radius, we can construct a probability map of circles for the image. We add this to regional property descriptors as the *circle probability*, indicating whether a region belongs to the circular color chart or not. Intermediate results of applying this method on a sample image are demonstrated in Fig. 4, in which, the constructed *centerMap* and circle probability map are able to correctly identify the color chart in the image border. Hitherto, 3 new shape features are added to regional property descriptors of mDRFI.

*Texture related features:* other than low-level texture descriptors like the variance of pixel values, DRFI uses the responses of Leung-Malik (LM) filters [30] to represent regions' texture properties. As illustrated in Fig. 5, LM filters mostly emphasize on tubular and spot like structures. To enhance and represent more complicated texture templates in the image, we propose to add the responses of Laws' filter bank [31] to the mDRFI as well. Laws' energy features emphasize on edge, spot, ripple, and wave structures in the texture of the image. There are 14 Laws' filters, whose responses can be used to extract textural features (14 new features). As one can see in the second row of Fig. 5, Laws' filters are able to target more complex texture structures than LM filters.

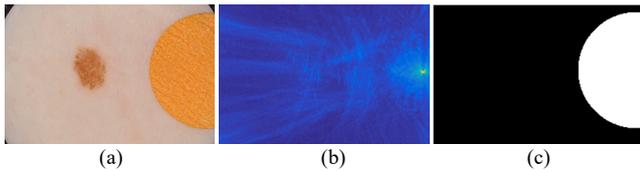

Fig. 4.   Procedure of circle detection for construction of the circle probability map: image (a) is the original input to the algorithm, (b) is the *centerMap* showing high values near the circle's center location, and (c) is obtained circle probability map (detected circle mask).

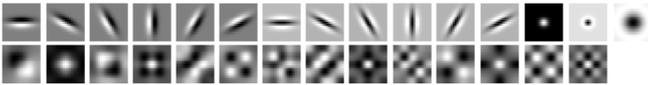

Fig. 5.   The top row shows 15 kernels of LM filter bank. The bottom row illustrates 14 Laws' filters which are able to enhance more complex templates.

In summary, we have added 3 shape (elongation, extent, and circle probability), 5 color (average of R, G, B, a*, and b*), and 14 texture (energy of Laws' filters responses) features to mDRFI. In addition to that, we propose to consider the segmentation level as an input feature to the random forest regressor as well. In total, we have extended the regional property descriptors of mDRFI by 23 new features listed in Table II with bold faces.

#### 1.2) New pseudo-background region

In dermoscopic images, healthy skin around the lesion can be assumed as background [10]. It is necessary to correctly specify the background region before any attempt to extract regional background descriptors. It is known that for a general dermoscopic dataset, such as ISBI2017, there are some images that include other objects (like color charts or dark corners) in their marginal areas. More importantly, some lesions extents reach the image borders. Therefore, those marginal regions should not be considered as background. To eliminate undesired effects of these problems, we introduce a new pseudo-background region.

First, we compute the histogram of the image and then smooth it by applying a moving average filter. The location of the last peak in the histogram is found and its 90th percentile is set as the background threshold to convert the image into a binary mask. Next, some post-processing operations are applied on the output background mask (filling holes, removing objects with area smaller than 500 pixels, and morphological closing using a disk-shaped structuring element of radius 5). Finally, a strip with 15 pixels width, starting from the outer boundary of the background object toward its inside, is selected as the pseudo-background region. Fig. 6 depicts the output of this procedure for a sample dermoscopic image that has both dark corners and a lesion touching the borders. It is desirable that the pseudo-background region only encompasses the healthy skin, as our proposed algorithm finds it in Fig. 6-(e).

#### 2) Thresholding the saliency map

The binary mask of the lesion is obtained by thresholding the output saliency map of the mDRFI method. A constant threshold of 0.5 has been used to convert the saliency map to binary mask. Of course, the binary mask has spurious parts in it which makes it unacceptable to be the lesion's initial mask. These are removed by steps taken below.

#### 3) Post processing (initial mask)

To make the binary mask more appropriate for the next segmentation part, we apply some post-processing tasks on it. First, an analysis of objects area should be carried out. The average area ($a_m$) and standard deviation ($a_s$) of objects' areas are calculated and then objects with areas smaller than $a_m - 2a_s$ are removed from the binary mask. The convex hull of the remaining objects is then constructed to serve as the initial mask of the lesion.

#### C. Final mask creation

The initial mask may not represent boundaries of the skin lesion perfectly. We propose to further refine the initial segmentation through level set evolution. We use the signed distance function of the initial mask as the initial state of the level set in a distance regularized level set evolution (DRLSE) framework [32], in order to drive it toward the lesion boundaries. Image channel with the highest entropy is chosen [6] to be used in DRLSE framework. After level set evolution, we apply some post-processing tasks (morphological opening and closing) on the DRLSE output to smooth out its boundaries and improve the results.

### IV. EXPERIMENTAL SETUP AND RESULTS

#### A. Datasets description

To evaluate our mDRFI approach and the proposed segmentation method, we use three different datasets. Two of these datasets are selected from international skin imaging collaboration (ISIC) archive and used in the challenge of "Skin Lesion Analysis toward Melanoma Detection" [26] held by the International Symposium on Biomedical Imaging (ISBI) in 2016 and 2017, which are abbreviated here as ISBI2016 and ISBI2017, respectively. Parts of the ISBI2016 dataset used in this article consisted of 900 and 397 images for training and testing, respectively. The ISBI2017 dataset, also, comprises 2000 images for training and 600 images as the testing set. Images in ISBI2016 and ISBI2017 have a great number of duplicate items. Therefore, we train and test our algorithm on each of these datasets separately. Another dataset that has been widely used in the field of dermoscopic image analysis is PH2 dataset [33] which comprises a total of 200 dermoscopic images. Prior to main processes, images are resized to a mutual size of 300×400 since they are captured with different resolution.

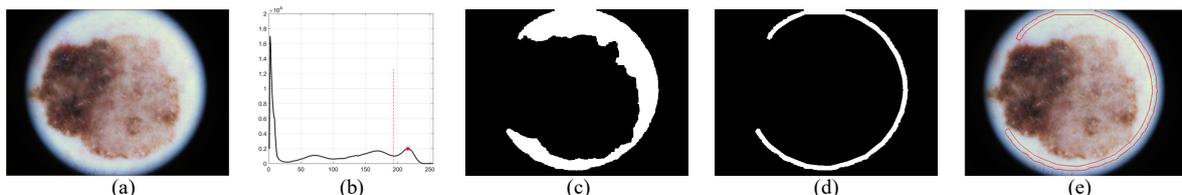

Fig. 6.   Obtaining new pseudo-background region: (a) original image, (b) histogram of the gray image, in which the last peak is found and the 90th percentile of its location is selected as the threshold, (c) thresholded and processed mask of background, (d) and (e) representing the new pseudo-background region.

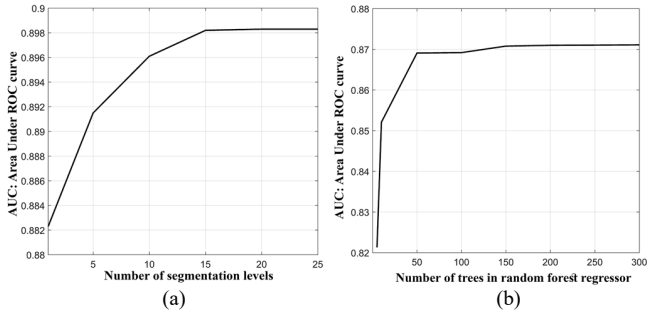

Fig. 7. Values of AUC for different configuration of mDRFI parameters.

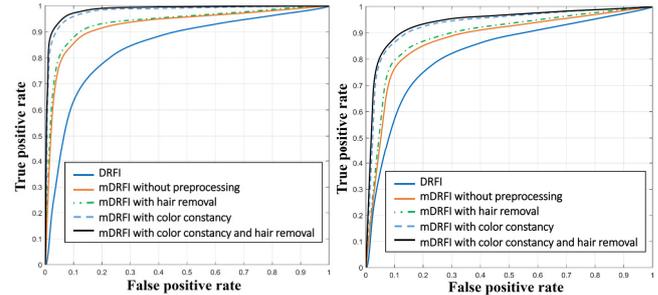

Fig. 8. ROC curves achieved by applying mDRFI (or DRFI) on ISBI2016 (left) and ISBI2017 (right) datasets with different preprocessing scenarios.

TABLE III
THE EFFECT OF DIFFERENT SALIENCY DETECTION AND PREPROCESSING SCENARIOS ON THE RESULTS OF THE PROPOSED SEGMENTATION ALGORITHM

| Saliency detection and preprocessing scenarios | Segmentation metrics | |
|---|---|---|
| | DSC | Acc |
| DRFI without preprocessing | 81.4 | 85.1 |
| DRFI with color constancy and hair removal | 84.8 | 88.9 |
| mDRFI without preprocessing | 85.9 | 89.5 |
| mDRFI with hair removal | 86.6 | 90.6 |
| mDRFI with color constancy | 89.5 | 92.8 |
| mdRFI with color constancy and hair removal | 90.7 | 94.3 |

### B. Implementation details

Codes are implemented in MATLAB R2014b, and all experiments are performed on an Intel Core i7-4790 machine with 16 GB memory running Ubuntu. Hair removal is done using GPU implementation of the Koehoorn algorithm[1] [28]. Our implementation of mDRFI is released[2], which consists of the implementation of image color constancy correction and all other novelties alongside the original DRFI source code. DRLSE toolbox for level set evolution is available online[3], as well.

### C. Selecting the optimal parameters for mDRFI

The most important parameters in the mDRFI framework that should be explored are the number of segmentation levels in the multi-level framework and the number of trees in the random forest regressor [21]. To find the optimal ranges for these parameters, we split the ISBI2017 training set into two subsets: a training subset with 1500 images and a validation subset comprising 500 images. First, we set the number of trees to a constant number of 150 and train mDRFI models with different numbers of segmentation levels on the training subset. Next, we use them to predict the saliency maps on the validation subset. The average values of AUC (area under ROC curves) obtained from the evaluation of 500 saliency maps are reported for each mDRFI configuration. As shown in Fig. 7-(a), increasing the number of segmentation levels elevates the mDRFI performance. This is because increasing the number of segmentation levels may form some regions that cover areas of the lesion with more confidence [21]. Based on this experiment and the AUC curve in Fig. 7-(a), we decide to set the number of segmentations to 15.

A similar experiment is carried out to assess the performance with changing the number of trees in the random forest regressor. It is observed that increasing the number of trees decreases the variances within the weak classifiers of the random forest regressor and lead to a better performance [21]. Considering the efficiency and computational load trade-off, we select the number of trees to be equal to 200. Variation of AUC performance score by changing the number of trees in random forest regressor is plotted in Fig. 7-(b).

Although we select the optimal parameters for training mDRFI, the plots in Fig. 7 show a small variation in AUC scores when the parameters are varied. For instance, by changing the number of segmentations from 1 to 25, the performance metric (AUC) varies only about 0.012 (1%). This implies that proposed mDRFI approach has a low sensitivity to its parameters.

### D. Exploring the effect of preprocessing

Different scenarios have been carried out to assess the effect of preprocessing on the saliency detection and final segmentation performances. In one scenario, we train the mDRFI model on the ISBI2016 training images and predict the saliency maps on the ISBI2016 test set, without applying any preprocessing on the images. In other scenarios, we apply hair removal, color constancy, or both. The outputs from each of these scenarios are then fed into the next steps of the proposed segmentation algorithm (thresholding, contour evolution, and post-processing). For the saliency detection, the ROC curves achieved for each of these scenarios are plotted in Fig. 8. Also, Table III reports the evaluation metrics for segmentation outputs from each preprocessing scenarios. Same processes are repeated for the ISBI2017 dataset and the results are depicted in the right side of Fig. 8 (the training and validation subsets for these experiments are similar to the subsets introduced in the previous subsection). Based on the ROC curves in Fig. 8 and segmentation results in Table III, we conclude that the mDRFI model trained with images that were processed with both hair removal and color constancy algorithms outperforms the other models. Therefore, this procedure is selected to be used in the proposed segmentation algorithm (results are reported in Tables IV-VII).

---

[1] Code provided at: http://www.cs.rug.nl/svcg/Shapes/HairRemoval
[2] Code released at: https://github.com/mjahanifar/mDRFI_matlab
[3] Code available at: http://www.imagecomputing.org/~cmli/DRLSE

*E. Quantitative and qualitative results*

To evaluate the segmentation algorithm quantitatively, we incorporate Dice Similarity Coefficient (DSC), Jaccard Similarity Index (JSI), and three pixel-wise performance metrics of Accuracy (Acc), Sensitivity (Sens), and Specificity (Spec) as described in [26]. The results of testing the proposed segmentation algorithm on different dermoscopic datasets are compared to the other state-of-the-art methods and reported in Tables IV-VII. It is noted that some evaluation metrics were absent for some methods. Thus, we replace their corresponding values by "-" in the tables.

When trained on the PH2 dataset, our algorithm outperforms the other methods in Table IV. PH2 is a relatively small and simple dataset that all of its images were captured using a fixed acquisition setup, and it is rational to achieve evaluation metrics as high as reported values in Table IV.

For ISBI2016 dataset, results are reported for both training and testing sets. As seen in Table V, our algorithm outperforms the other saliency detection based algorithms (Fan *et al.* [22] and Ahn *et al.* [23]) and our previous method in Zamani *et al.* [4] on the ISBI2016 train set. Its performance is slightly worse than that of Yuan *et al.* [15], which is based on deep convolutional neural networks (DCNNs). Evaluation of the ISBI2016 test set is also reported in Table VI. Based on this table, our algorithm is listed within top 3 algorithms, outperforming other methods that rank 2 to 5 at the ISBI2016 Part 1: Segmentation Challenge.

In the same way, we apply our method on the ISBI2017 test set and present the results in Table VII. Our proposed algorithm ranked seventh among 21 teams competing in the "Part 1: Lesion Segmentation" of the "ISBI 2017 Skin Lesion Analysis Towards Melanoma Detection" with insignificant differences from the other superior methods. The algorithm proposed by Yuan *et al.* [15] ranked first place in this challenge as well. Our proposed algorithm achieves average Dice value of 0.839 and Jaccard index of 0.749 on 600 images of the ISBI2017 test set, which is only 0.010 and 0.015, respectively, less than Dice and Jaccard values reported as the first ranked method (Yading Yuan). The differences from other superior competitors in Table VII are even smaller (other evaluation metrics have the same trend). These results prove that the proposed method can achieve a good segmentation performance, comparable to the other state-of-the-art methods, which mostly are based on DCNNs. Note that the ISBI2016 and ISBI2017 training sets were used to train the saliency maps used in Table V/VI, and Table VII, respectively.

In order to assess the proposed segmentation algorithm qualitatively, we deploy it on several dermoscopic images (in two categories of simple and extreme cases). Fig. 9 shows the segmentation results and their corresponding saliency maps (the middle row). The green contours represent the ground truth segmentation, red dashed lines stand for the contour of the lesion's initial mask (constructed by thresholding the saliency map as explained in section III-B), and the blue contours show the borders of the refined final segmentation through level set evolution (as described in section III-C).

TABLE IV
RESULTS OF PH2 DATASET SEGMENTATIONS

| Method | Average of evaluation metrics (%) | | | | |
|---|---|---|---|---|---|
| | DSC | JSI | Acc | Sens | Spec |
| Pennisi [11] | - | - | 89.4 | 71.0 | 97.1 |
| Barata [7] | 90.0 | 83.7 | 92.8 | 90.4 | 97.0 |
| Ahn [23] | 91.5 | - | - | - | - |
| Fan [22] | 89.3 | - | 93.6 | 87.0 | - |
| Zamani [4] | 92.0 | 85.8 | 96.5 | 95.4 | 98.1 |
| Proposed | 95.2 | 92.3 | 97.9 | 97.2 | 98.9 |

TABLE V
RESULTS OF ISBI2016 TRAIN SET SEGMENTATIONS

| Method | Average of evaluation metrics (%) | | | | |
|---|---|---|---|---|---|
| | DSC | JSI | Acc | Sens | Spec |
| Fan [22] | 81.8 | - | 91.8 | 74.7 | - |
| Ahn [23] | 83.9 | - | - | - | - |
| Zamani [4] | 89.5 | 80.2 | 93.5 | 83.2 | 98.7 |
| Proposed | 91.7 | 85.5 | 94.2 | 88.7 | 98.4 |

TABLE VI
RESULTS OF ISBI2016 TEST SET SEGMENTATIONS*

| Method | Average of evaluation metrics (%) | | | | |
|---|---|---|---|---|---|
| | DSC | JSI | Acc | Sens | Spec |
| Yuan [15] | 91.2 | 87.4 | 95.5 | 91.8 | 96.6 |
| Proposed | 90.7 | 83.8 | 94.3 | 90.1 | 98.2 |
| 1) EXB | 91.0 | 84.3 | 95.3 | 91.0 | 96.5 |
| 2) CUMED (Yu [5]) | 89.7 | 82.9 | 94.9 | 91.1 | 95.7 |
| 3) Mahmudur | 89.5 | 82.2 | 95.2 | 88.0 | 96.9 |
| 4) SFUmial | 88.5 | 81.1 | 94.4 | 91.5 | 95.5 |
| 5) TMUteam (Zamani [4]) | 88.8 | 81.0 | 94.6 | 83.2 | 98.7 |

\* Numbered methods indicate rankings directly adopted from ISBI2016 challenge: https://challenge.kitware.com/#phase/566744dccad3a56fac786787

TABLE VII
RESULTS OF ISBI2017 TEST SET SEGMENTATIONS**

| Method | Average of evaluation metrics (%) | | | | |
|---|---|---|---|---|---|
| | DSC | JSI | Acc | Sens | Spec |
| Yading Yuan | 84.9 | 76.5 | 93.4 | 82.5 | 97.5 |
| Matt Berseth | 84.7 | 76.2 | 93.2 | 82.0 | 97.8 |
| Lei Bi | 84.4 | 76.0 | 93.4 | 80.2 | 98.5 |
| Euijoon Ahn | 84.2 | 75.8 | 93.4 | 80.1 | 98.4 |
| RECOD titans | 83.9 | 75.4 | 93.1 | 81.7 | 96.9 |
| Jeremy Kawahara | 83.7 | 75.2 | 93.0 | 81.3 | 97.6 |
| Jahanifar Zamani (Proposed) | 83.9 | 74.9 | 93.0 | 81.0 | 98.1 |

\*\*Methods are adopted and listed according to rankings in the ISBI2017 challenge: https://challenge.kitware.com/#phase/584b0afacad3a51cc66c8e24

V. DISCUSSION

To demonstrate the advantages of the mDRFI and its newly added features (regional property and pseudo-background descriptors introduced in section III-B), we trained the DRFI on the ISBI2016 train set and applied it on the ISBI2016 test set as well. Based on the ROC curves for saliency detection in Fig. 8 and evaluation metrics for segmentation in Table III, it is obvious that mDRFI outperforms the DRFI method by a large margin. Without applying any preprocessing on the data, the segmentation outputs using the proposed algorithm with the mDRFI and DRFI derived saliency maps achieve DSC scores of 85.9% and 81.4%, respectively. This large improvement is due to the newly introduced features in the proposed mDRFI framework.

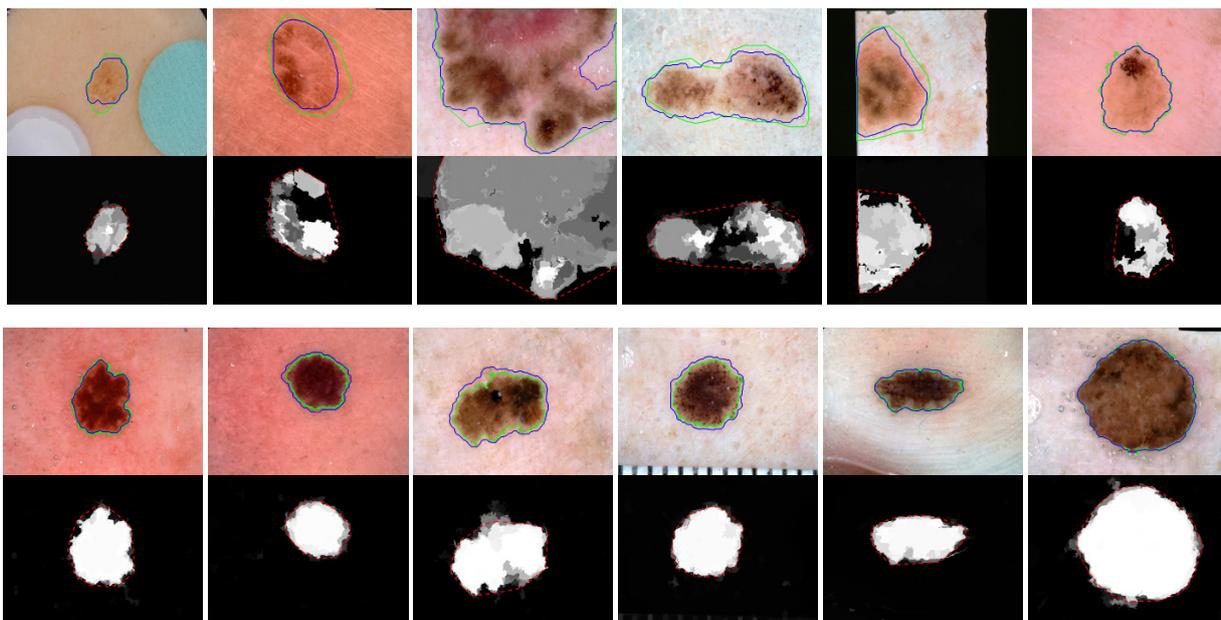

Fig. 9.   Qualitative assessment of the proposed segmentation method. The first and the second row illustrates six example of extreme cases with their corresponding saliency maps through mDRFI. The third and the fourth rows depict that information for six relatively simple cases. In all images the green, blue, and red (dashed) contours correspond to the borders of ground truth, final, and initial segmentations (convex hull of the thresholded saliency map), respectively.

Another important observation is that the color constancy preprocessing is far more effective than hair removal in improving the segmentation results. As seen in Fig. 8 and Table III, the proposed mDRFI with hair removal performs slightly better than the mDRFI without preprocessing. However, it performs considerably better on images preprocessed for color constancy. It also can be observed that hair removal marginally improves the segmentation performance achieved using only color constancy. This is due to the fact that mDRFI is able to detect the hair-like structures by itself by using regional property descriptors that deal with elongation. On the other hand, adding color constancy normalizes shades of colors in all images, leading to a better training of mDRFI.

Deep convolutional neural networks (DCNNs) are very powerful tools that conquered the first ranks in all parts of the ISBI 2016 and ISBI 2017 challenges. DCNNs are neural network architectures comprising stacks of convolutional, fully connected, nonlinearity (like Relu), max pooling, and often deconvolutional layers [5], [34]. The function of convolutional/deconvolutional layers in DCNNs is to extract visual features from patches of the image to help network decide if a particular patch belongs to the lesion area or not. The basic idea of the mDRFI model is the same as the DCNNs, extracting features from different regions of the image and deciding whether it belongs to the lesion or not. Furthermore, likewise DCNNs with pooling and unpooling layers, mDRFI implements a multi-level segmentation mechanism. However, the difference between DCNNs and mDRFI is that the first group learns the features (convolution filters) automatically through the learning phase, in the latter, features are hand-crafted. The advantage of DCNNs is that they can learn higher level features as the network goes deeper and the number of convolutions increases. But in the mDRFI model, we only have features that are introduced earlier. Thus, DCNNs are able to outperform mDRFI. It is worth mentioning that unlike the classification applications that may need deep and high-level features, there is no need for very high-level features in the segmentation applications [34]. That is why our proposed mDRFI can achieve a close performance to the most of the state-of-the-art DCNNs (see the small difference between mDRFI performance and other DCNN based methods in Tables IV-VII).

However, there is a drawback to using DCNNs. They need large training sets to be trained well enough and be able to learn high-level features. For the case of mDRFI, mid-level features are hand-crafted for the specific problem of dermoscopic image segmentation. Therefore, mDRFI can easily be trained on small datasets and outperform the other training based, or saliency-based methods, as presented in Table IV for the PH2 dataset.

## VI.  Conclusion

In this paper, we modified a supervised saliency detection algorithm to perform better on dermoscopic images. We used the proposed saliency detection algorithm to obtain an initial mask of the lesion and then evolved it in a level set framework to achieve the final segmentation. Our method was tested on three well-known dermoscopic datasets. The proposed algorithm outperformed all other state-of-the-art methods that have been published on the PH2 dataset. For more general datasets like ISBI2016 and ISBI2017, our algorithm performed close to the most powerful deep learning based approaches.

## REFERENCES

[1] Anonymous, "Skin Cancers," *World Health Organization*, 2017. [Online]. Available: http://www.who.int/uv/faq/skincancer/en/index1.html. [Accessed: 25-Aug-2017].

[2] B. W. Stewart and C. P. Wild, *World Cancer Report 2014*. World Health Organization, 2014, p. 953.

[3] N. C. F. Codella, Q.-B. Nguyen, S. Pankanti, D. A. Gutman, B. Helba, A. C. Halpern, and J. R. Smith, "Deep learning ensembles for melanoma recognition in dermoscopy images," *IBM J. Res. Dev.*, vol. 61, no. 4, pp. 1–5, 2017.

[4] N. Zamani Tajeddin and B. Mohammadzadeh Asl, "A General Algorithm for Automatic Lesion Segmentation in Dermoscopy Images," in *23rd Iranian Conference on Biomedical Engineering and 2016 1st International Iranian Conference on Biomedical Engineering (ICBME)*, 2016, no. November, pp. 134–139.

[5] L. Yu, H. Chen, Q. Dou, J. Qin, and P.-A. Heng, "Automated melanoma recognition in dermoscopy images via very deep residual networks," *IEEE Trans. Med. Imaging*, vol. 36, no. 4, pp. 994–1004, 2017.

[6] M. Silveira, J. C. Nascimento, J. S. Marques, A. R. S. Marçal, T. Mendonça, S. Yamauchi, J. Maeda, and J. Rozeira, "Comparison of segmentation methods for melanoma diagnosis in dermoscopy images," *Sel. Top. Signal Process. IEEE J.*, vol. 3, no. 1, pp. 35–45, 2009.

[7] C. Barata, M. Ruela, M. Francisco, T. Mendonça, and J. S. Marques, "Two Systems for the Detection of Melanomas in Dermoscopy Images Using Texture and Color Features," *IEEE Syst. J.*, vol. 8, no. 3, pp. 965–979, 2013.

[8] M. E. Celebi, Q. Wen, S. Hwang, H. Iyatomi, and G. Schaefer, "Lesion Border Detection in Dermoscopy Images Using Ensembles of Thresholding Methods," *Ski. Res. Technol.*, vol. 19, no. 1, pp. e252–e258, Dec. 2013.

[9] M. Emre Celebi, H. A. Kingravi, H. Iyatomi, Y. Alp Aslandogan, W. V. Stoecker, R. H. Moss, J. M. Malters, J. M. Grichnik, A. A. Marghoob, H. S. Rabinovitz, and S. W. Menzies, "Border detection in dermoscopy images using statistical region merging," *Ski. Res. Technol.*, vol. 14, no. 3, pp. 347–353, 2008.

[10] H. Zhou, G. Schaefer, A. H. Sadka, and M. E. Celebi, "Anisotropic mean shift based fuzzy c-means segmentation of dermoscopy images," *Sel. Top. Signal Process. IEEE J.*, vol. 3, no. 1, pp. 26–34, 2009.

[11] A. Pennisi, D. D. Bloisi, D. Nardi, A. R. Giampetruzzi, C. Mondino, and A. Facchiano, "Skin lesion image segmentation using Delaunay Triangulation for melanoma detection," *Comput. Med. Imaging Graph.*, vol. 52, pp. 89–103, 2016.

[12] H. Zhou, X. Li, G. Schaefer, M. E. Celebi, and P. Miller, "Mean shift based gradient vector flow for image segmentation," *Comput. Vis. Image Underst.*, vol. 117, no. 9, pp. 1004–1016, 2013.

[13] H. Iyatomi, H. Oka, M. E. Celebi, M. Hashimoto, M. Hagiwara, M. Tanaka, and K. Ogawa, "An improved internet-based melanoma screening system with dermatologist-like tumor area extraction algorithm," *Comput. Med. Imaging Graph.*, vol. 32, no. 7, pp. 566–579, 2008.

[14] F. Peruch, F. Bogo, M. Bonazza, V.-M. Cappelleri, and E. Peserico, "Simpler, faster, more accurate melanocytic lesion segmentation through meds," *IEEE Trans. Biomed. Eng.*, vol. 61, no. 2, pp. 557–565, 2014.

[15] Y. Yuan, M. Chao, and Y. C. Lo, "Automatic Skin Lesion Segmentation Using Deep Fully Convolutional Networks with Jaccard Distance," *IEEE Trans. Med. Imaging*, vol. 36, no. 9, pp. 1876–1886, 2017.

[16] M. E. Celebi, Q. Wen, H. Iyatomi, K. Shimizu, H. Zhou, and G. Schaefer, "A State-of-the-Art Survey on Lesion Border Detection in Dermoscopy Images," in *Dermoscopy Image Analysis*, M. E. Celebi, T. Mendonca, and J. S. Marques, Eds. CRC Press, 2015, pp. 97–129.

[17] R. B. Oliveira, E. Mercedes Filho, Z. Ma, J. P. Papa, A. S. Pereira, and J. M. R. S. Tavares, "Computational methods for the image segmentation of pigmented skin lesions: a review," *Comput. Methods Programs Biomed.*, vol. 131, pp. 127–141, 2016.

[18] N. K. Mishra and M. E. Celebi, "An Overview of Melanoma Detection in Dermoscopy Images Using Image Processing and Machine Learning," *arXiv Prepr. arXiv1601.07843*, pp. 1–15, 2016.

[19] A. Borji, "What is a salient object? A dataset and a baseline model for salient object detection," *IEEE Trans. Image Process.*, vol. 24, no. 2, pp. 742–756, 2015.

[20] A. Borji, M.-M. Cheng, H. Jiang, and J. Li, "Salient object detection: A benchmark," *IEEE Trans. Image Process.*, vol. 24, no. 12, pp. 5706–5722, 2015.

[21] J. Wang, H. Jiang, Z. Yuan, M.-M. Cheng, X. Hu, and N. Zheng, "Salient Object Detection: A Discriminative Regional Feature Integration Approach," *Int. J. Comput. Vis.*, vol. 123, no. 2, pp. 251–268, 2017.

[22] H. Fan, F. Xie, Y. Li, Z. Jiang, and J. Liu, "Automatic segmentation of dermoscopy images using saliency combined with Otsu threshold," *Comput. Biol. Med.*, vol. 85, pp. 75–85, 2017.

[23] E. Ahn, J. Kim, L. Bi, A. Kumar, C. Li, M. Fulham, and D. D. Feng, "Saliency-based Lesion Segmentation via Background Detection in Dermoscopic Images," *IEEE J. Biomed. Heal. informatics*, vol. 21, no. 6, pp. 1685–1693, 2017.

[24] P. F. Felzenszwalb and D. P. Huttenlocher, "Efficient graph-based image segmentation," *Int. J. Comput. Vis.*, vol. 59, no. 2, pp. 167–181, 2004.

[25] L. Breiman and A. Cutler, "Breiman and Cutler's Random Forests for Classification and Regression," 2015. [Online]. Available: https://www.stat.berkeley.edu/~breiman/RandomForests/.

[26] N. C. F. Codella, D. Gutman, M. E. Celebi, B. Helba, M. A. Marchetti, S. W. Dusza, A. Kalloo, K. Liopyris, N. Mishra, H. Kittler, and A. Halpern, "Skin Lesion Analysis Toward Melanoma Detection: A Challenge at the 2017 International Symposium on Biomedical Imaging (ISBI), Hosted by the International Skin Imaging Collaboration (ISIC)," *arXiv: 1710.05006 [cs.CV]*, 2017.

[27] C. Barata, M. E. Celebi, and J. S. Marques, "Improving dermoscopy image classification using color constancy," *IEEE J. Biomed. Heal. informatics*, vol. 19, no. 3, pp. 1146–1152, 2015.

[28] J. Koehoorn, A. C. Sobiecki, D. Boda, A. Diaconeasa, S. Doshi, S. Paisey, A. Jalba, and A. Telea, "Automated digital hair removal by threshold decomposition and morphological analysis," in *International Symposium on Mathematical Morphology and Its Applications to Signal and Image Processing*, 2015, vol. 9082, pp. 15–26.

[29] S. Basalamah, "Histogram based circle detection," *Int. J. Comput. Sci. Netw. Secur.*, vol. 12, no. 8, pp. 40–43, 2012.

[30] T. Leung and J. Malik, "Representing and recognizing the visual appearance of materials using three-dimensional textons," *Int. J. Comput. Vis.*, vol. 43, no. 1, pp. 29–44, 2001.

[31] K. I. Laws, "Rapid texture identification," in *24th annual technical symposium*, 1980, pp. 376–381.

[32] C. Li, C. Xu, C. Gui, and M. D. Fox, "Distance regularized level set evolution and its application to image segmentation," *IEEE Trans. image Process.*, vol. 19, no. 12, pp. 3243–3254, 2010.

[33] T. Mendonça, P. M. Ferreira, J. S. Marques, A. R. S. Marcal, and J. Rozeira, "PH2 - A dermoscopic image database for research and benchmarking," in *35th Annual International Conference of the IEEE Engineering in Medicine and Biology Society (EMBC)*, 2013, pp. 5437–5440.

[34] G. Litjens, T. Kooi, B. E. Bejnordi, A. A. A. Setio, F. Ciompi, M. Ghafoorian, J. A. W. M. van der Laak, B. van Ginneken, and C. I. Sánchez, "A survey on deep learning in medical image analysis," *Med. Image Anal.*, vol. 42, pp. 60–88, 2017.